\pdfoutput=1
\documentclass[11pt,a4paper]{article}
\usepackage[hyperref]{acl2021}
\usepackage{times}
\usepackage{latexsym}
\usepackage{tabularx}
\usepackage{graphicx}
\usepackage{multirow}
\usepackage{booktabs}
\usepackage{float}

\usepackage{microtype}

\aclfinalcopy 
\def\aclpaperid{25} 

\definecolor{cadmiumorange}{rgb}{0.93, 0.53, 0.18}
\definecolor{darkspringgreen}{rgb}{0.09, 0.45, 0.27}
\definecolor{chamoisee}{rgb}{0.63, 0.47, 0.35}

\title{Realistic Evaluation Principles \\for Cross-document Coreference Resolution}

\author{Arie Cattan\textsuperscript{1} \quad
    Alon Eirew\textsuperscript{1,2} \quad
    Gabriel Stanovsky\textsuperscript{3} \quad
    Mandar Joshi\textsuperscript{4} \quad
    Ido Dagan\textsuperscript{1}\\ \\
\textsuperscript{1}Computer Science Department, Bar Ilan University \\ 
\textsuperscript{2}Intel Labs, Israel \quad
\textsuperscript{3}The Hebrew University of Jerusalem 
 \\
\textsuperscript{4}Allen School of Computer Science \& Engineering, University of Washington, Seattle, WA
 \\
  {\tt  arie.cattan@gmail.com} \quad {\tt  alon.eirew@intel.com} \\ 
  {\tt gabis@cse.huji.ac.il} \quad
     {\tt mandar90@cs.washington.edu} \\ {\tt  dagan@cs.biu.ac.il}  
  }

\date{}

\begin{document}
\maketitle

\begin{abstract}
We point out that common evaluation practices for cross-document coreference resolution have been unrealistically permissive in their assumed settings, yielding inflated results. We propose addressing this issue via two evaluation methodology principles. First, as in other tasks, models should be evaluated on predicted mentions rather than on gold mentions. Doing this raises a subtle issue regarding singleton coreference clusters, which we address by decoupling the evaluation of mention detection from that of coreference linking. Second, we argue that models should not exploit the synthetic topic structure of the standard ECB+ dataset, forcing models to confront the lexical ambiguity challenge, as intended by the dataset creators. We demonstrate empirically the drastic impact of our more realistic evaluation principles on a competitive model, yielding a score which is 33 F1 lower compared to evaluating by prior lenient practices.\footnote{\url{https://github.com/ariecattan/coref}} 
\end{abstract}
\section{Introduction}
\label{sec:intro}

Cross-document (CD) coreference resolution identifies and links textual mentions that refer to the same entity or event across multiple documents. For example, Table~\ref{tab:subtopic} depicts different news stories involving former U.S. president Barack Obama. 

While subsuming the challenges of within-document~(WD) coreference, CD coreference introduces additional unique challenges. Most notably, lexical similarity is often not a good indicator when identifying cross-document links, as documents are authored independently. 
As shown in Table~\ref{tab:subtopic}, the same event can be referenced using different expressions (``nominated'', ``approached"), while two \emph{different} events can be referenced using the same expression (``name'').
Despite these challenges, reported state-of-the-art results on the popular CD coreference ECB+ benchmark~\cite{cybulska-vossen-2014-using} are relatively high, reaching up to 80 F1~\citep{barhom-etal-2019-revisiting, meged-etal-2020-paraphrasing}.

\begin{table}
    
    \centering
    \resizebox{0.48\textwidth}{!}{
    \begin{tabular}{p{9cm}}
    \toprule
    \multicolumn{1}{c}{\textbf{Subtopic 1}} \\
    \toprule

    \textbf{Doc 1:} \emph{News that \textbf{\textcolor{blue}{Barack Obama}} may \textbf{\underline{\textcolor{cadmiumorange}{name}}} \textbf{\textcolor{red}{Dr. Sanjay Gupta}} of Emory University and CNN as \textbf{\textcolor{blue}{his}} Surgeon...} \\
    \textbf{Doc 2:} \emph{CNN's management confirmed yesterday that \textbf{\textcolor{red}{Dr. Gupta}} had been  \textbf{\textcolor{cadmiumorange}{\underline{approached}}} by the Obama team. }\\
    \toprule
    \multicolumn{1}{c}{\textbf{Subtopic 2}} \\
    \toprule

    \textbf{Doc 3:} \emph{\textbf{\textcolor{blue}{President Obama}} will \textbf{\underline{\textcolor{darkspringgreen}{name}}} \textbf{\textcolor{chamoisee}{Dr. Regina Benjamin}} as Surgeon General in a Rose Garden announcement...} \\

   \textbf{Doc 4:} \emph{\textbf{\textcolor{blue}{Obama}} \textbf{\underline{\textcolor{darkspringgreen}{nominates}}} new surgeon general: \textbf{\textcolor{chamoisee}{genius grant fellow Dr. Benjamin. }} \textbf{\textcolor{blue}{He}} emphasizes \textbf{\textcolor{blue}{his}} \textbf{\underline{\textcolor{darkspringgreen}{decision}}}..} 
    \\
    
    \bottomrule
    \end{tabular}}
    \caption{Example of sentences of from the ECB+. The underlined words represent events, same color represents a coreference cluster.
    Different documents describe the same event using different words (e.g name, approached), while the two predicates ``name" in the two subtopics are \emph{not} coreferring. 
    }
    \label{tab:subtopic}
\end{table}

In this paper, we show that CD coreference models achieve these numbers using overly-permissive evaluation protocols, namely assuming gold entity and event mentions are given, rewarding singletons and bypassing the lexical ambiguity challenge. Accordingly, we present more realistic evaluation principles which better reflect model performance in real-world scenarios.

First, following well established standards in WD coreference resolution~\citep{pradhan-etal-2012-conll}, we propose that CD coreference models should be also evaluated on predicted mentions. While recent models unrealistically assume that event mentions are given as part of the input, practical application on new texts and domains requires performing coreference on raw text, including automatic mention detection. 
Using predicted mentions raises a subtle point with regards to singletons (entities which are only referenced once).
In particular, we observe that   ECB+'s inclusion of singletons inaccurately rewards models for predicting them, by conflating the evaluation of mention identification with that of coreference detection. To address this, we propose reporting of singleton identification performance in a separate metric, while reporting coreference results without singletons.

Second, we find that ECB+ does not accurately reflect real-world scenarios where prominent events can be referenced in documents spanning different subjects and domains. To facilitate its annotation, ECB+ mimics this phenomenon by artificially grouping documents dealing with the same event (e.g., the nomination of Sanjay Gupta in Table~\ref{tab:subtopic}) into a \emph{subtopic}, and further groups two similar subtopics into a larger \emph{topic} document group (e.g., different nominations of government officials in Table~\ref{tab:subtopic}). We observe that recent works exploit ECB+'s artificially simplistic structure by practically running the coreference model at the subtopic level, thus sidestepping a major lexical ambiguity challenge (e.g., mentions of ``nomination'' across subtopics do not co-refer). In contrast, in real-world scenarios such clustering is much harder to perform and is often not as easily delineated. For example, Barack Obama and events from his presidency can be referenced in news, literature, sport reports, and more. To address this, we propose that models report performance also at the topic level.

Finally, we show empirically that both of these evaluation practices artificially inflate results. An  end-to end model that outperforms state-of-the-art results on previous evaluation settings drops by 33 F1 points when using our proposed evaluation scheme, pointing at weaknesses that future modelling work could explore.

\section{Background}
\label{sec:bg}
In this work, we will examine the evaluation of CD coreference on the popular ECB+ corpus~\citep{cybulska-vossen-2014-using}, constructed as an augmentation of the EECB and ECB datasets~\citep{lee-etal-2012-joint, bejan-harabagiu-2010-unsupervised}.
As exemplified in Table~\ref{tab:subtopic}, ECB+ groups its annotated documents into \textit{subtopics}, consisting of different reports of the same real-world event (e.g., the nomination of Sanjay Gupta), and \emph{topics}, which in turn consist of two lexically similar subtopics. Full ECB+ details are presented in Appendix~\ref{app:dataset}.

The ECB+ evaluation protocol largely follows that of CoNLL-2012, perhaps the most popular WD benchmark~\citep{pradhan-etal-2012-conll}, with two major distinctions.
First, barring a few notable exceptions~\citep{yang-etal-2015-hierarchical, choubey-huang-2017-event},\footnote{However, as noted in~\citep{barhom-etal-2019-revisiting}, they consider only the intersection between gold and predicted mentions, not penalizing models for false positive mention identification.} most recent CD models have unrealistically assumed that gold entity and event mentions are given as part of the input, reducing the task to finding coreference links between gold mentions~\citep{Bejan2014UnsupervisedEC, cybulska-vossen-2015-translating, kenyon-dean-etal-2018-resolving, barhom-etal-2019-revisiting, meged-etal-2020-paraphrasing}. Second, while singletons are omitted on CoNLL-2012, they are exhaustively annotated in ECB+.

In the following section, we present a more realistic evaluation framework for CD coreference, taking into account the interacting distinctions of ECB+.

\section{Realistic Evaluation Principles}

In this paper, we suggest that CD coreference models should perform and be evaluated on predicted mentions. 
To achieve this, in Section~\ref{subsec:singleton_effect}, we will introduce the \emph{singleton effect} on coreference evaluation and propose to decouple the evaluation of mention prediction from coreference resolution. In Section~\ref{subsec:topic_level}, we will establish guidelines allowing to better assess how models handle the ubiquitous lexical ambiguity challenge in real-world scenarios.

\subsection{Decoupling Coreference Evaluation}
\label{subsec:singleton_effect}

Our goal is to propose a more reliable evaluation methodology of a coreference system over predicted mentions when singletons are included.

We use an example to show that evaluating singleton prediction with standard coreference metrics (B3, CEAF, LEA) could lead to counterproductive results which are hard to interpret (henceforth, we refer to this phenomenon as the \emph{singleton effect}). 
Assume $G$ denotes the gold clusters for Table~\ref{tab:subtopic} (for brevity, we omit some mentions), and $S1$ and $S2$ denote the output of two systems, which differ in their mention detection and coreference link performance:\footnote{This follows the natural distribution of singletons (about 50\%), as illustrated in PreCo~\citep{chen-etal-2018-preco}.}

\vspace{-7pt}
\begin{table}[H]
    \centering
    \resizebox{0.48\textwidth}{!}{
    \begin{tabular}{lp{80mm}}
        $G$ & \{News\}, \{Emory University\}, \{confirmed\}, \{yesterday\}, \{announcement\}, \textcolor{cadmiumorange}{\{name, approached\}}, \textcolor{darkspringgreen}{\{names, nominates, decision\}}\\
        $S1$ & \{News\}, \{Emory University\}, \{confirmed\}, \{yesterday\}, \{announcement, \textcolor{cadmiumorange}{name, approached}, \textcolor{darkspringgreen}{names, nominates, decision}\}\\
        $S2$ & \{News that\}, \{Emory\}, \{announcement, \textcolor{cadmiumorange}{name, approached}\}, \textcolor{darkspringgreen}{\{names, nominates, decision\}} 
    \end{tabular}}
    \label{tab:my_label}
\end{table}
\vspace{-12pt}

$S1$ identified the mentions of the singleton clusters while $S2$ missed them and predicted incorrect span boundaries for the two first mentions (``News that" and ``Emory"). Both $S1$ and $S2$ erroneously merged the singleton mention ``announcement" with the cluster \textcolor{cadmiumorange}{\{name, approached\}}; however, $S1$ further included these mentions with the lexically-similar cluster \textcolor{darkspringgreen}{\{names, nominates, decision\}}, whereas $S2$ successfully separated them. In other words, $S1$ performs well on the mention detection task, but worse on the coreference linking, and $S2$ did the opposite.

\begin{table}[!t]
    \centering
    \resizebox{0.48\textwidth}{!}{
    \begin{tabular}{lccccccc}
    \toprule
    \phantom{fwidsv} && MUC & B\textsuperscript{3} & CEAFe & LEA & CoNLL\\
    \midrule
    \multirow{2}{*}{CoNLL-2012} & $S1$ & 75.0 & 53.1 & 44.4 & 42.1 & 57.5 \\
    & $S2$ & \textbf{85.7} & \textbf{83.9} & \textbf{90.0} & \textbf{80.0} & \textbf{86.5} \\
    \midrule
    \multirow{2}{*}{With Singletons} & $S1$ & 75.0 & \textbf{77.6} & \textbf{77.8} & \textbf{69.0} & \textbf{76.8} \\
    & $S2$ & \textbf{85.7} & 59.2 & 32.7 & 50.0 & 59.2 \\
    \bottomrule
    \end{tabular}}
    \caption{Coreference results of $S1$ and $S2$ with (1) the standard CoNLL-2012 evaluation, where $S2$ does better and (2) when including singletons, where $S1$ does better. $S2$ predicts the coreference links better than $S1$ but $S1$ achieves higher results in (2) because $S1$ performs better the mention detection task.}
    \label{tab:singletons}
\end{table}

Table~\ref{tab:singletons} shows the results of $S1$ and $S2$ according to (1) the common CoNLL-2012 evaluation, where only non-singleton clusters are evaluated, and (2) using coreference metrics also on singleton prediction.
With respect to (1), $S2$ achieves higher results according to all evaluation metrics. In (2), we see the opposite, the results of $S1$ are significantly higher than $S2$ w.r.t B\textsuperscript{3} (+18.4), CEAF-e (+45.1), and LEA (+19), but not w.r.t MUC, a link-based metric. Indeed, these evaluation metrics reward $S1$ in both recall and precision for all predicted singletons, while penalizing $S2$ for the wrong and missing singleton spans. Since singletons are abundant in natural text, they contribute greatly to the overall score. 
However, as observed by~\citet{rahman-ng-2009-supervised}, a model's ability to identify that these singletons do not belong to any coreference cluster is already captured in the evaluation metrics, and additional penalty is not desired. 
In Appendix~\ref{app:singleton}, we introduce the aforementioned evaluation metrics for coreference resolution (MUC, B\textsuperscript{3}, CEAF and LEA) and explain how singletons affect them.


\begin{table*}[!ht]
    \centering
    \resizebox{\textwidth}{!}{
    \begin{tabular}{@{}clccccccccccccccclc@{}}
    \toprule
    && \multicolumn{3}{c}{MUC} && \multicolumn{3}{c}{$B^3$} & & \multicolumn{3}{c}{$CEAFe$} && \multicolumn{3}{c}{LEA} && CoNLL\\
    \cmidrule{3-5} \cmidrule{7-9} \cmidrule{11-13} \cmidrule{15-17} \cmidrule{19-19}
    && R & P & $F_1$ && R & P & $F_1$ && R &P & $F_1$ && R &P & $F_1$ && $F_1$  \\ 
   \midrule
   \multirow{10}{*}{\shortstack{Subtopic \\Clustering}} &
        Singleton baseline$^+$ & 0 & 0 & 0 && 45.2 & 100 & 62.3 &&  86.7 & 39.2 & 54.0 && 35.0 & 35.0 & 35.0 && 38.8 \\
        & Singleton baseline$^-$ & 0 & 0 & 0 && 0 & 0 & 0 && 0 & 0 & 0 &&0&0&0 && 0\\
        
        & \citet{barhom-etal-2019-revisiting}$^+$ & 78.1 & 84.0 & 80.9 && 76.8 & 86.1 & 81.2 && 79.6 & 73.3 & 76.3 && 64.6 & 72.3 & 68.3 && 79.5  \\
        &\citet{barhom-etal-2019-revisiting}$^-$ & 78.1 & 84.0 & 80.9 &&  61.2 & 73.5 & 66.8 && 63.2 & 48.9 & 55.2 && 58.4 & 71.2 & 64.2 && 67.6  \\
        
        &\citet{meged-etal-2020-paraphrasing}$^+$ & 78.8 & 84.7 & 81.6 && 75.9 & 85.9 & 80.6 && 81.1 & 74.8 & \textbf{77.8} && 64.7 & 73.4 & 68.8 && 80.0  \\
        &\citet{meged-etal-2020-paraphrasing}$^-$ & 78.8 & 84.7 & 81.6 && 60.4 & 73.8 & 66.4 && 65.5 & 49.5 & 56.4 && 57.2 & 71.2 & 63.4 && 68.1 \\
        
        \cmidrule{2-19}
        &Our model \textendash{} Gold$^+$  &  85.1 & 81.9 & \textbf{83.5} && 82.1 & 82.7 & \textbf{82.4} && 75.2 & 78.9 & 77.0 && 68.8 & 72.0 & \textbf{70.4} && \textbf{81.0}   \\
        &Our model \textendash{} Gold$^-$ & 85.1 & 81.9 & 83.5 && 70.8 & 70.2 & 70.5 && 68.2 & 52.3 & 59.2 && 68.2 & 67.6 & 67.9 && 71.1   \\
        &Our model \textendash{} Predicted$^+$  &  61.7 & 67.4 & 64.5 && 57.8 & 68.4 & 62.6 && 57.2 & 65.5 & 61.1 && 46.6 & 57.7 & 51.6 && 62.7   \\
        &Our model \textendash{} Predicted$^-$ & 61.7 & 67.4 & 64.5 && 47.6 & 56.9 & 51.8 && 53.0 & 41.9 & 46.8 && 44.4 & 53.8 & 48.7 && 54.4   \\

    \midrule
    \multirow{4}{*}{Topic Level} & Our model \textendash{} Gold$^+$ & 80.1 & 76.3 & \textbf{78.1} && 77.4 & 71.7 & \textbf{74.5} && 73.1 & 77.8 & \textbf{75.4} && 62.9 & 59.1 & \textbf{61.0} && \textbf{76.0} \\ 
     & Our model \textendash{} Gold$^-$ & 80.1 & 76.3 & 78.1 && 63.4 & 54.1 & 58.4 && 56.3 & 44.2 & 49.5 && 59.7 & 49.6 & 54.2 && 62.0 \\
    &Our model \textendash{} Predicted$^+$ & 61.5 & 62.5 & 62.0 && 55.6 & 56.1 & 55.8 && 52.8 & 66.7 & 59.0 && 43.4 & 46.2 & 44.8 && 58.9 \\
    &Our model \textendash{} Predicted$^-$ & 61.5 & 62.5 & 62.0 && 44.7 & 41.4 & 43.0 && 43.9 & 37.9 & 40.7  && 40.9 & 37.4 & 39.1 && 48.6 \\
    \bottomrule
    \end{tabular}}
    \caption{Event coreference on ECB+ test, while including($^+$)/excluding($^-$) singletons in the evaluation, showing that (1) including singletons in coreference metrics inflate performance in all models, (2) using predicted mentions (see rows marked ``Predicted'') over gold mentions harms performance, (3) topic level evaluation (bottom part) is markedly lower than subtopic performance, showing that models struggle with lexical ambiguity, and 
    (4) our model outperforms previous models on most F1 scores (see numbers in bold).}
    \label{tab:subtopic_results_event}
\end{table*}

To address the \emph{singleton effect}, we suggest decoupling the evaluation of the two coreference substasks, mention detection and coreference linking, allowing to better analyze coreference results and to compare systems more appropriately.\footnote{This also makes possible to compare coreference results across datasets that include/omit singletons, addressing an issue raised by~\citet{stoyanov-etal-2009-conundrums}.}

Mention detection is typically a span detection task and should be evaluated using standard span metrics on all detected mentions, including singletons. In particular, we use the span F1 metric and consider a predicted mention as correct if it has an exact match with a gold mention, as common in named entity recognition~\citep{tjong-kim-sang-de-meulder-2003-introduction}. Using such evaluation in our above example, $S1$ achieves 100 F1 and $S2$ achieves 66.7 F1 (recall: 60, precision: 75).

For the coreference evaluation, we propose to follow CoNLL-2012 and apply coreference metrics only on non-singleton (gold and predicted) clusters, as singletons are already evaluated under the mention detection evaluation. We note also that even when omitting singletons, coreference metrics still penalize models for making coreference errors involving singletons (as $S2$ is penalized for linking ``announcement" to a cluster). 

We further show empirically~(§\ref{subsec:results}) that when evaluating using gold mentions, the \emph{singleton effect} is amplified and harms the validity of the current CD evaluation protocol. Evidently, a dummy baseline that predicts no coreference links and puts each input gold mention in a singleton cluster achieves non-negligible performance~\cite{luo-2005-coreference}, while state-of-the-art results are artificially inflated.

\subsection{Confronting Lexical Ambiguity}
\label{subsec:topic_level}

As mentioned previously, the same event can be described in documents from different topics, while documents in the same topic may describe \emph{different} events (e.g. different nominations as surgeon general, as shown in Table~\ref{tab:subtopic}). Such settings pose a lexical ambiguity problem, where models encounter identical or lexically-similar words that should be assigned to different coreference clusters. Accordingly, while topical document clustering is useful for CD coreference resolution in general, it does not solve the ambiguity problem and models still need to make subtle disambiguation distinctions (e.g nomination of Sanjay Gupta vs. nomination of Regina Benjamin). Aiming at simulating this challenge on a manageable annotation task, the ECB+ authors~\citep{cybulska-vossen-2014-using} augmented each topic in the original ECB with an additional subtopic of the same event type, allowing to challenge models with lexical ambiguity~(as mentioned in Section~\ref{sec:bg}).

However, recent works~\citep{barhom-etal-2019-revisiting, meged-etal-2020-paraphrasing} predict coreference clusters separately on each subtopic, using a simple unsupervised document clustering during preprocessing. Such clustering performs near perfectly on ECB+ because of its synthetic structure, where each topic includes exactly two subtopics with only a few coreference links across different subtopics. Yet, document clustering is not expected to perform as well in realistic settings where coreferring events can spread multiple topics. More importantly, this bypasses intentions behind the inclusion of subtopics in the ECB+'s and avoids challenging the coreference models on lexical ambiguity. Indeed, the ECB+ authors, in a subsequent work, did not apply a topic clustering~\cite{cybulska-vossen-2015-translating}. 

We therefore recommend that models report results also at the \emph{topic} level (when document clustering is not applied). 
This will conform to ECB+'s purpose and follows the original evaluation setup of the ECB+ corpus~\citep{Bejan2014UnsupervisedEC}.   
\section{Experiments}
\label{sec:results}

We show empirically that each of the previous evaluation practices (using gold mentions, singleton inclusion, and subtopic clustering) artificially inflates the results~(§\ref{subsec:results}). As recent CD coreference models are designed to perform on gold mentions~(§\ref{sec:bg}), we cannot use them to set baseline results on predicted mentions. We therefore develop a simple and efficient end-to-end model for CD coreference resolution by combining the successful single document \emph{e2e-coref}~\citep{lee-etal-2017-end} with common CD modeling approaches.

\subsection{Model}
\label{subsec:model}

We briefly describe the general architecture of our model, further details are explained in~\citep{Cattan2021CrossdocumentCR} and Appendix~\ref{app:model}.
Given a set of documents, our model operates in four sequential steps: (1) following~\citet{lee-etal-2017-end}, we encode all possible spans up to a length $n$ with the concatenation of four vectors: the output representations of the span boundary (first and last) tokens, an attention-weighted sum of token representations in the span, and a feature vector denoting the span length (2) we train a mention detector on the ECB+ mentions, and keep further spans with a positive score,\footnote{Here, we deviate from~\citet{Cattan2021CrossdocumentCR} who dynamically prune spans during training, because we need to predict singleton clusters.} (3) we generate positive and negative coreference pairs on the predicted mentions and train a pairwise scorer, and (4) apply an agglomerative clustering on the pairwise similarity scores to form the coreference clusters at inference.

\subsection{Results}
\label{subsec:results}

We first evaluate our model under the current evaluation setup (gold mentions, singletons, subtopic) and compare it with two recent neural state-of-the-art models~\cite{barhom-etal-2019-revisiting,meged-etal-2020-paraphrasing}.
In addition, we test a dummy \emph{singleton} baseline which puts each gold mention in a singleton cluster and re-evaluate all baselines while omitting singletons.
The results in Table~\ref{tab:subtopic_results_event} show that our model surpasses current state-of-the-art results in previous settings, supporting its relevance for setting baseline results over predicted mentions. The mention detection performance of our model is 80.1 F1 (Recall 76 and Precision 84.7).

The results corroborate the importance of our proposed evaluation enhancements.
First, the performance drops dramatically when using predicted mentions (e.g. from 71.1 to 54.4 F1 at the subtopic level).
Second, for all models, the results are significantly higher when including singletons in coreference metrics, because, as explained in Section~\ref{subsec:singleton_effect}, models are rewarded for singleton prediction. Indeed, the model performs better in mention detection than in coreference linking, confirming the importance of decoupling the evaluation of the two subtasks.
Finally, performance is lower at the \emph{topic} level than at the \emph{subtopic} level (62.0 vs. 71.1 F1 using gold mentions and 48.6 vs. 54.4 F1 using predicted mentions), indicating that models struggle with lexical ambiguity~(§\ref{subsec:topic_level}).
Taken together, evaluating over raw text without singletons while not clustering into fine-grained subtopics, leads to a performance drop of 33 F1 points, indicating the vast room for improvement under realistic settings.

\section{Conclusion}

We established two realistic evaluation principles for CD coreference resolution: (1) predicting mentions and (2) facing the lexical ambiguity challenge. We also set baseline results for future work on our evaluation methodology using a SOTA model.
\section*{Acknowledgment}

We thank Shany Barhom for fruitful discussion and sharing code, and Yehudit Meged for providing her coreference predictions. The work described herein was supported in part by grants from Intel Labs, Facebook, the Israel Science Foundation grant 1951/17, the Israeli Ministry of Science and Technology, the German Research Foundation through the German-Israeli Project Cooperation (DIP, grant DA 1600/1-1), and from the Allen Institute for AI.
\section*{Ethical Considerations}


\paragraph{Model}  As described in the supplementary material~(§\ref{app:model}), our cross-document coreference model does not contain any intentional biasing or ethical issues, and our experiments were conducted on a single 12GB GPU, with relatively low compute time.
\bibliographystyle{acl_natbib}
\bibliography{anthology, acl2021}

\begin{thebibliography}{23}
\expandafter\ifx\csname natexlab\endcsname\relax\def\natexlab#1{#1}\fi

\bibitem[{Bagga and Baldwin(1998)}]{bagga-baldwin-1998-entity-based}
Amit Bagga and Breck Baldwin. 1998.
\newblock \href {https://doi.org/10.3115/980845.980859} {Entity-based
  cross-document coreferencing using the vector space model}.
\newblock In \emph{36th Annual Meeting of the Association for Computational
  Linguistics and 17th International Conference on Computational Linguistics,
  Volume 1}, pages 79--85, Montreal, Quebec, Canada. Association for
  Computational Linguistics.

\bibitem[{Barhom et~al.(2019)Barhom, Shwartz, Eirew, Bugert, Reimers, and
  Dagan}]{barhom-etal-2019-revisiting}
Shany Barhom, Vered Shwartz, Alon Eirew, Michael Bugert, Nils Reimers, and Ido
  Dagan. 2019.
\newblock \href {https://doi.org/10.18653/v1/P19-1409} {Revisiting joint
  modeling of cross-document entity and event coreference resolution}.
\newblock In \emph{Proceedings of the 57th Annual Meeting of the Association
  for Computational Linguistics}, pages 4179--4189, Florence, Italy.
  Association for Computational Linguistics.

\bibitem[{Bejan and Harabagiu(2014)}]{Bejan2014UnsupervisedEC}
C.~Bejan and Sanda~M. Harabagiu. 2014.
\newblock Unsupervised event coreference resolution.
\newblock \emph{Computational Linguistics}, 40:311--347.

\bibitem[{Bejan and Harabagiu(2010)}]{bejan-harabagiu-2010-unsupervised}
Cosmin Bejan and Sanda Harabagiu. 2010.
\newblock \href {https://www.aclweb.org/anthology/P10-1143} {Unsupervised event
  coreference resolution with rich linguistic features}.
\newblock In \emph{Proceedings of the 48th Annual Meeting of the Association
  for Computational Linguistics}, pages 1412--1422, Uppsala, Sweden.
  Association for Computational Linguistics.

\bibitem[{Cattan et~al.(2021)Cattan, Eirew, Stanovsky, Joshi, and
  Dagan}]{Cattan2021CrossdocumentCR}
Arie Cattan, Alon Eirew, Gabriel Stanovsky, Mandar Joshi, and Ido Dagan. 2021.
\newblock Cross-document coreference resolution over predicted mentions.
\newblock In \emph{Findings of the Association for Computational Linguistics:
  ACL 2021}, Online. Association for Computational Linguistics.

\bibitem[{Chen et~al.(2018)Chen, Fan, Lu, Yuille, and
  Rong}]{chen-etal-2018-preco}
Hong Chen, Zhenhua Fan, Hao Lu, Alan Yuille, and Shu Rong. 2018.
\newblock \href {https://doi.org/10.18653/v1/D18-1016} {{P}re{C}o: A
  large-scale dataset in preschool vocabulary for coreference resolution}.
\newblock In \emph{Proceedings of the 2018 Conference on Empirical Methods in
  Natural Language Processing}, pages 172--181, Brussels, Belgium. Association
  for Computational Linguistics.

\bibitem[{Choubey and Huang(2017)}]{choubey-huang-2017-event}
Prafulla~Kumar Choubey and Ruihong Huang. 2017.
\newblock \href {https://doi.org/10.18653/v1/D17-1226} {Event coreference
  resolution by iteratively unfolding inter-dependencies among events}.
\newblock In \emph{Proceedings of the 2017 Conference on Empirical Methods in
  Natural Language Processing}, pages 2124--2133, Copenhagen, Denmark.
  Association for Computational Linguistics.

\bibitem[{Cybulska and Vossen(2014)}]{cybulska-vossen-2014-using}
Agata Cybulska and Piek Vossen. 2014.
\newblock \href
  {http://www.lrec-conf.org/proceedings/lrec2014/pdf/840_Paper.pdf} {Using a
  sledgehammer to crack a nut? lexical diversity and event coreference
  resolution}.
\newblock In \emph{Proceedings of the Ninth International Conference on
  Language Resources and Evaluation ({LREC}'14)}, pages 4545--4552, Reykjavik,
  Iceland. European Language Resources Association (ELRA).

\bibitem[{Cybulska and Vossen(2015)}]{cybulska-vossen-2015-translating}
Agata Cybulska and Piek Vossen. 2015.
\newblock \href {https://doi.org/10.3115/v1/W15-0801} {Translating granularity
  of event slots into features for event coreference resolution.}
\newblock In \emph{Proceedings of the The 3rd Workshop on {EVENTS}: Definition,
  Detection, Coreference, and Representation}, pages 1--10, Denver, Colorado.
  Association for Computational Linguistics.

\bibitem[{Joshi et~al.(2019)Joshi, Levy, Zettlemoyer, and
  Weld}]{joshi-etal-2019-bert}
Mandar Joshi, Omer Levy, Luke Zettlemoyer, and Daniel Weld. 2019.
\newblock \href {https://doi.org/10.18653/v1/D19-1588} {{BERT} for coreference
  resolution: Baselines and analysis}.
\newblock In \emph{Proceedings of the 2019 Conference on Empirical Methods in
  Natural Language Processing and the 9th International Joint Conference on
  Natural Language Processing (EMNLP-IJCNLP)}, pages 5803--5808, Hong Kong,
  China. Association for Computational Linguistics.

\bibitem[{Kenyon-Dean et~al.(2018)Kenyon-Dean, Cheung, and
  Precup}]{kenyon-dean-etal-2018-resolving}
Kian Kenyon-Dean, Jackie Chi~Kit Cheung, and Doina Precup. 2018.
\newblock \href {https://doi.org/10.18653/v1/S18-2001} {Resolving event
  coreference with supervised representation learning and clustering-oriented
  regularization}.
\newblock In \emph{Proceedings of the Seventh Joint Conference on Lexical and
  Computational Semantics}, pages 1--10, New Orleans, Louisiana. Association
  for Computational Linguistics.

\bibitem[{Lee et~al.(2012)Lee, Recasens, Chang, Surdeanu, and
  Jurafsky}]{lee-etal-2012-joint}
Heeyoung Lee, Marta Recasens, Angel Chang, Mihai Surdeanu, and Dan Jurafsky.
  2012.
\newblock \href {https://www.aclweb.org/anthology/D12-1045} {Joint entity and
  event coreference resolution across documents}.
\newblock In \emph{Proceedings of the 2012 Joint Conference on Empirical
  Methods in Natural Language Processing and Computational Natural Language
  Learning}, pages 489--500, Jeju Island, Korea. Association for Computational
  Linguistics.

\bibitem[{Lee et~al.(2017)Lee, He, Lewis, and Zettlemoyer}]{lee-etal-2017-end}
Kenton Lee, Luheng He, Mike Lewis, and Luke Zettlemoyer. 2017.
\newblock \href {https://doi.org/10.18653/v1/D17-1018} {End-to-end neural
  coreference resolution}.
\newblock In \emph{Proceedings of the 2017 Conference on Empirical Methods in
  Natural Language Processing}, pages 188--197, Copenhagen, Denmark.
  Association for Computational Linguistics.

\bibitem[{Liu et~al.(2019)Liu, Ott, Goyal, Du, Joshi, Chen, Levy, Lewis,
  Zettlemoyer, and Stoyanov}]{liu2019roberta}
Yinhan Liu, Myle Ott, Naman Goyal, Jingfei Du, Mandar Joshi, Danqi Chen, Omer
  Levy, Mike Lewis, Luke Zettlemoyer, and Veselin Stoyanov. 2019.
\newblock Ro{BERT}a: A robustly optimized {BERT} pretraining approach.
\newblock \emph{arXiv preprint arXiv:1907.11692}.

\bibitem[{Luo(2005)}]{luo-2005-coreference}
Xiaoqiang Luo. 2005.
\newblock \href {https://www.aclweb.org/anthology/H05-1004} {On coreference
  resolution performance metrics}.
\newblock In \emph{Proceedings of Human Language Technology Conference and
  Conference on Empirical Methods in Natural Language Processing}, pages
  25--32, Vancouver, British Columbia, Canada. Association for Computational
  Linguistics.

\bibitem[{Meged et~al.(2020)Meged, Caciularu, Shwartz, and
  Dagan}]{meged-etal-2020-paraphrasing}
Yehudit Meged, Avi Caciularu, Vered Shwartz, and Ido Dagan. 2020.
\newblock \href {https://doi.org/10.18653/v1/2020.findings-emnlp.440}
  {Paraphrasing vs coreferring: Two sides of the same coin}.
\newblock In \emph{Findings of the Association for Computational Linguistics:
  EMNLP 2020}, pages 4897--4907, Online. Association for Computational
  Linguistics.

\bibitem[{Moosavi and Strube(2016)}]{moosavi-strube-2016-coreference}
Nafise~Sadat Moosavi and Michael Strube. 2016.
\newblock \href {https://doi.org/10.18653/v1/P16-1060} {Which coreference
  evaluation metric do you trust? a proposal for a link-based entity aware
  metric}.
\newblock In \emph{Proceedings of the 54th Annual Meeting of the Association
  for Computational Linguistics (Volume 1: Long Papers)}, pages 632--642,
  Berlin, Germany. Association for Computational Linguistics.

\bibitem[{Pradhan et~al.(2012)Pradhan, Moschitti, Xue, Uryupina, and
  Zhang}]{pradhan-etal-2012-conll}
Sameer Pradhan, Alessandro Moschitti, Nianwen Xue, Olga Uryupina, and Yuchen
  Zhang. 2012.
\newblock \href {https://www.aclweb.org/anthology/W12-4501} {{C}o{NLL}-2012
  shared task: Modeling multilingual unrestricted coreference in
  {O}nto{N}otes}.
\newblock In \emph{Joint Conference on {EMNLP} and {C}o{NLL} - Shared Task},
  pages 1--40, Jeju Island, Korea. Association for Computational Linguistics.

\bibitem[{Rahman and Ng(2009)}]{rahman-ng-2009-supervised}
Altaf Rahman and Vincent Ng. 2009.
\newblock \href {https://www.aclweb.org/anthology/D09-1101} {Supervised models
  for coreference resolution}.
\newblock In \emph{Proceedings of the 2009 Conference on Empirical Methods in
  Natural Language Processing}, pages 968--977, Singapore. Association for
  Computational Linguistics.

\bibitem[{Stoyanov et~al.(2009)Stoyanov, Gilbert, Cardie, and
  Riloff}]{stoyanov-etal-2009-conundrums}
Veselin Stoyanov, Nathan Gilbert, Claire Cardie, and Ellen Riloff. 2009.
\newblock \href {https://www.aclweb.org/anthology/P09-1074} {Conundrums in noun
  phrase coreference resolution: Making sense of the state-of-the-art}.
\newblock In \emph{Proceedings of the Joint Conference of the 47th Annual
  Meeting of the {ACL} and the 4th International Joint Conference on Natural
  Language Processing of the {AFNLP}}, pages 656--664, Suntec, Singapore.
  Association for Computational Linguistics.

\bibitem[{Tjong Kim~Sang and
  De~Meulder(2003)}]{tjong-kim-sang-de-meulder-2003-introduction}
Erik~F. Tjong Kim~Sang and Fien De~Meulder. 2003.
\newblock \href {https://www.aclweb.org/anthology/W03-0419} {Introduction to
  the {C}o{NLL}-2003 shared task: Language-independent named entity
  recognition}.
\newblock In \emph{Proceedings of the Seventh Conference on Natural Language
  Learning at {HLT}-{NAACL} 2003}, pages 142--147.

\bibitem[{Vilain et~al.(1995)Vilain, Burger, Aberdeen, Connolly, and
  Hirschman}]{vilain-etal-1995-model}
Marc Vilain, John Burger, John Aberdeen, Dennis Connolly, and Lynette
  Hirschman. 1995.
\newblock \href {https://www.aclweb.org/anthology/M95-1005} {A model-theoretic
  coreference scoring scheme}.
\newblock In \emph{Sixth Message Understanding Conference ({MUC}-6):
  Proceedings of a Conference Held in {C}olumbia, {M}aryland, November 6-8,
  1995}.

\bibitem[{Yang et~al.(2015)Yang, Cardie, and
  Frazier}]{yang-etal-2015-hierarchical}
Bishan Yang, Claire Cardie, and Peter Frazier. 2015.
\newblock \href {https://doi.org/10.1162/tacl_a_00155} {A hierarchical
  distance-dependent {B}ayesian model for event coreference resolution}.
\newblock \emph{Transactions of the Association for Computational Linguistics},
  3:517--528.

\end{thebibliography}

\clearpage
\appendix

\section{The ECB+ Dataset}
\label{app:dataset}
Documents in ECB+ were selected from various topics in the Google News archive in English, while annotation was performed separately for each topic. 
ECB+ statistics are shown in Table~\ref{tab:ecb_stat}. 
As opposed to Ontonotes, only a few sentences are exhaustively annotated in each document, and the annotations include singletons. 

In addition, it is worth noting that the ECB+ authors kept the entities from EECB~\citep{lee-etal-2012-joint} only if they participate in events in the annotated sentences, while leaving all other entities. Accordingly, ``Los Angeles" and ``Los Angeles hospital" are marked as coreferent in the sentences \emph{``Yesterday in \textbf{\underline{Los Angeles}}, pin-up icon Bettie Page succumbed to complications..} and \emph{''Pinup icon Bettie Page died Thursday evening at \textbf{\underline{a hospital in Los Angeles..}}"} because they refer to the location of the same event. This differs from the standard entity coreference resolution since detecting those entities involves an additional challenge of extracting event participants, for example, using a Semantic Role Labeling system. 

\begin{table}[!h]
    \centering
    \resizebox{0.48\textwidth}{!}{
    \begin{tabular}{@{}llll@{}}
    \toprule
    & \textbf{Train} & \textbf{Validation} & \textbf{Test} \\
    \midrule
    \# Topics & 25 & 8 & 10 \\
    \# Documents & 594 & 196 & 206 \\
    \# Sentences & 1037 & 346 & 457 \\
    \# Mentions & 3808/4758 & 1245/1476 & 1780/2055 \\
    \# Singletons & 1116/814 & 280/205 & 632/412 \\
    \# Clusters & 411/472 & 129/125 & 182/196 \\
    \bottomrule
    \end{tabular}}
    \caption{ECB+ statistics. \# Clusters do not include singletons. The slash numbers for \# Mentions, \# Singletons, and \# Clusters represent event/entity statistics. As recommended by the authors in the release note, we follow the split of \citet{cybulska-vossen-2015-translating} that uses a curated subset of the dataset.}
    \label{tab:ecb_stat}
\end{table}

\section{Singleton Effect on Coreference Metrics}
\label{app:singleton}
Here, we briefly introduce the different evaluation metrics for coreference resolution (MUC, B\textsuperscript{3}, CEAF and LEA) and explain how singletons affect them. As mentioned in the paper, all evaluation metrics penalize models for wrongly linking a singleton to a cluster or singletons together. However, B3, CEAF and LEA further reward models for predicting singleton clusters, as explained below.

\paragraph{MUC} 
Introduced by~\cite{vilain-etal-1995-model}, MUC is an early link-based evaluation metric for coreference resolution. Recall and precision are measured based on the minimal number of coreference links needed to align gold and predicted clusters, as follows:
\begin{equation}
    Recall = \frac{\sum_{k_{i} \in K} (|k_{i}| - |p(k_i)|)}{\sum_{k_{j} \in K} (|k_{j}| - 1)}
\end{equation}
where $p(k_i)$ is the set of different predicted clusters that contain one or more mention of the gold cluster $k_i$. The precision is obtained by switching the role of the predicted and the gold clusters. Since MUC scores are calculated over the coreference links, singletons do not affect this metric, as observed in our illustrative example in the paper (Section~\ref{subsec:singleton_effect}).

\paragraph{B\textsuperscript{3}} 
B\textsuperscript{3}~\citep{bagga-baldwin-1998-entity-based} is a mention-based evaluation metric, the recall and precision correspond to the average of individual mention scores. The recall is defined as the proportion of its true coreferering mentions that the system links, over all the gold coreferering mentions that are linked to it, as follows:
\begin{equation}
    \label{eq:b3}
    Recall(m_i) = \frac{|Rm_i \cap Km_i|}{|Km_i|} 
\end{equation}
where $Rm_i$ and $Km_i$ are respectively the system and the gold cluster containing the mention $m_i$. The precision is obtained by switching the role of the predicted and gold clusters.

Here, all mentions $m_i$ (including singleton mentions) are scored in Eq.~\ref{eq:b3} and participate in the overall recall and precision score. Therefore, a singleton that was successfully predicted will be rewarded 100\% in both precision and recall, missing singletons will affect the recall and extra-singletons will affect the precision. 

\paragraph{CEAF} 
Introduced by~\citet{luo-2005-coreference}, CEAF assumes that each predicted cluster should be mapped to only one gold cluster and vice versa. 
Using the Kuhn-Munkres algorithm, CEAF first finds the best one-to-one mapping $g(*)$ of the predicted clusters to the gold clusters, according to a similarity function $\phi$. Given this mapping, predicted clusters are compared to their corresponding gold clusters, as follows:
\begin{equation}
    Recall = \frac{\sum_{r_i \in R} \phi (r_i, g^*(r_i)) }{\sum_{k_i \in K}\phi (k_i, k_i)}
\end{equation}
where $R$ is the set of predicted clusters, $K$ the set of gold clusters, $g^*(r_i)$ the gold cluster aligned to the predicted cluster $r_i$, and $\phi()$ the similarity function. The precision is obtained by switching the role of the predicted and gold clusters in the denominator.  There are two variants of CEAF based on $\phi$, (1) a mention-based CEAFm defined as the number of shared mentions between the two clusters $\phi(r_i, k_i) = |r_i \cap k_i|$ and (2) an entity-based metric CEAFe: $\phi(r_i, k_i) = 2 \frac{|r_i \cap k_i|}{|r_i|+ |k_i|}$. 
Here again, a predicted singleton cluster that appears also in the gold will be obviously mapped to it and will be rewarded 100\% in both recall and precision.

\paragraph{LEA} 
Recently proposed by~\citet{moosavi-strube-2016-coreference}, LEA is the most recent evaluation metric, designed to overcome shortcomings in previous evaluation metrics, notably the \emph{mention identification effect} in B\textsuperscript{3} and CEAF. LEA is a Link-Based Entity-Aware metric, which assigns a score to each coreference cluster, based on all coreference links ($n \times (n-1) / 2$) in the cluster, as follows:
\begin{equation}
    Recall = \frac{\sum_{k_i \in K} (|k_{i}| \times \sum_{r_j \in R} \frac{link(k_i \cap r_j)}{link(k_i)}) }{\sum_{k_z \in K}|k_{z}|}
\end{equation}
where $link(k_i)$ is the total number of links in the gold cluster $k_i$, $link(k_i, r_j)$ is the total number of links in the predicted cluster $r_j$ that appears in the gold cluster $k_i$, and $|k_i|$ is the number of mentions in the gold cluster $k_i$ in order to give higher importance to large clusters. The precision is calculated by switching the role of the gold clusters $K$ and the predicted clusters $R$. 
Singleton clusters are also rewarded because they have self-links (links to themselves). However, since each cluster score is weighted by the size of the cluster, the \emph{singleton effect} is less important in LEA, as we can see in the paper~(Table~\ref{tab:subtopic_results_event}).

\section{Our Coreference Model}
\label{app:model}

As mentioned in the paper~(§\ref{subsec:model}), our model is inspired by the single document coreference resolver \emph{e2e-coref}~\citep{lee-etal-2017-end}. The \emph{e2e-coref} model forms the coreference clusters by linking each mention to an antecedent span appearing before it in the text. However, in the CD setting, there is no linear ordering between the documents. We therefore implement a new model while modifying the clustering method and the optimization function of the original \emph{e2e-coref} model, as elaborated below.\footnote{Please refer to~\citet{Cattan2021CrossdocumentCR} for more details, results and ablations of the model.} 

\paragraph{Span Representation}
Given a set of documents, the first step consists of encoding each document separately using RoBERTa\textsubscript{\emph{LARGE}}~\cite{liu2019roberta}. Long documents are split into non overlapping segments of up to 512 word-piece tokens and are encoded independently \cite{joshi-etal-2019-bert}. We then, following~\citet{lee-etal-2017-end}, represent each possible span up to a length $n$ with the concatenation of four vectors: the output representations of the span boundary (first and last) tokens, an attention-weighted sum of token representations in the span, and a feature vector denoting the span length. We use $g_i$ to refer to the vector representation of the span $i$.

\paragraph{Mention Scorer}
We train a mention detector $s_m(i)$ using a simple MLP on top of these span representations, indicating whether $i$ is a mention in ECB+. This is possible because singleton mentions are annotated in ECB+~(§\ref{app:dataset}). Unlike the \emph{e2e-coref}, we keep further only detected mentions in both training and inference. We also tried the joint approach but the performance drops by 0.4 CoNLL F1 and the run-time was longer.

\paragraph{Pairwise Scorer}
Given the predicted mentions, we first generate positive and negative training pairs as follows. The positive instances consist of all the pairs of mentions that belong to the same coreference cluster, while the negative examples are sampled (20x the number of positive pairs) from all other pairs. This sampling reduces the computation time, and limits the unbalanced negative ratio between training pairs.
Then, for each pair of mentions $i$ and $j$, we concatenate 3 vectors: $g_i$, $g_j$, and the element-wise multiplication $g_i \circ g_j$, and feed it to a simple MLP, which outputs a score $s(i, j)$ indicating the likelihood that mentions $i$ and $j$ belong to the same cluster, which we optimize using the binary cross-entropy loss on the pair label. Due to memory constraints, we freeze output representations from RoBERTa instead of fine-tuning all parameters.

\paragraph{Agglomerative Clustering}
As common in recent CD coreference models~\citep{yang-etal-2015-hierarchical, choubey-huang-2017-event, kenyon-dean-etal-2018-resolving, barhom-etal-2019-revisiting, meged-etal-2020-paraphrasing}, we use an agglomerative clustering on the pairwise scores $s(i, j)$ to form the coreference clusters at inference time. 
The agglomerative clustering step merges the most similar cluster pairs until their pairwise similarity score falls below a tuned threshold $\tau$.

\paragraph{Technical Details} We conduct our experience on a single GeForce GTX 1080 Ti 12GB GPU. Our model has 14M parameters. On average, the training takes 30 minutes and inference over all the test set takes 3 minutes.

\end{document}